\documentclass[10pt,final,conference,letterpaper,oneside,twocolumn]{IEEEtran}

\usepackage{amsmath}
\usepackage{amsfonts}
\usepackage{amssymb}
\usepackage{cite}
\usepackage{graphicx}
\usepackage{multirow}
\usepackage{algorithm}
\usepackage{algorithmicx}
\usepackage{algpseudocode}
\usepackage{booktabs}
\usepackage{flushend}

\algrenewcommand\algorithmicrequire{\textbf{Input:}}
\algrenewcommand\algorithmicensure{\textbf{Output:}}

\ifCLASSOPTIONcompsoc
	\usepackage[caption=false,font=normalsize,labelfont=sf,textfont=sf]{subfig}
\else
	\usepackage[caption=false,font=footnotesize]{subfig}
\fi

\usepackage{blindtext}

\title{Multi-atomic Annealing Heuristic for~Static~Dial-a-ride~Problem}

\author{
	\IEEEauthorblockN{
		Song Guang Ho\IEEEauthorrefmark{1},
		Ramesh Ramasamy Pandi\IEEEauthorrefmark{2}, 
		Sarat Chandra Nagavarapu\IEEEauthorrefmark{3} and 
		Justin Dauwels\IEEEauthorrefmark{4}}
	\IEEEauthorblockA{ %\IEEEauthorrefmark{1}
		School of Electrical and Electronic Engineering\\
		Nanyang Technological University, 50 Nanyang Avenue, Singapore, 639798.}
	\IEEEauthorblockA{
		Email: 
		\IEEEauthorrefmark{1}hosongguang@gmail.com, 
		\IEEEauthorrefmark{2}ramesh006@e.ntu.edu.sg,
		\IEEEauthorrefmark{3}saratchandra.nagavarapu@gmail.com,
		\IEEEauthorrefmark{4}jdauwels@ntu.edu.sg}
		}

\begin{document}

\maketitle

\begin{abstract}
Dial-a-ride problem (DARP) deals with the transportation of users between pickup and drop-off locations associated with specified time windows. This paper proposes a novel algorithm called multi-atomic annealing (MATA) to solve static dial-a-ride problem. Two new local search operators (\textit{burn} and \textit{reform}), a new construction heuristic and two request sequencing mechanisms (\textit{Sorted\_List} and \textit{Random\_List}) are developed. Computational experiments conducted on various standard DARP test instances prove that MATA is an expeditious meta-heuristic in contrast to other existing methods. In all experiments, MATA demonstrates the capability to obtain high quality solutions, faster convergence, and quicker attainment of a first feasible solution. It is observed that MATA attains a first feasible solution 29.8 to 65.1\% faster, and obtains a final solution that is 3.9 to 5.2\% better, when compared to other algorithms within 60 sec.
\end{abstract}


\section{Introduction}
Dial-a-ride problem belongs to a specific class of vehicle routing problem \cite{VRP}, in which passengers request for transportation between specified origin and destination. Each passenger usually defines time window for either pickup or drop-off.

DARP has been studied extensively since it was introduced by Wilson \cite{1} in 1971; many attempts have been made to develop solution methodologies for DARP. Due to the high complexity of DARP, solvers generally require a long time to converge and obtain high quality solution, causing them impractical in real life, especially in a dynamic environment where changes happen frequently (i.e., requests, vehicles and traffic conditions). Exact methods such as branch-and-cut \cite{4} have been developed, which require a long execution time especially for large-scale problems. Consequently, many heuristics and meta-heuristics have emerged to overcome this issue. In past decade, various techniques such as simulated annealing \cite{2} and tabu search \cite{TS}\cite{7} have been proposed to achieve near-optimal solutions in a shorter time.

An evolutionary local search approach \cite{5} has been proposed, in which a dynamic probabilities management mechanism is used in the local search to improve convergence. A competitive variable neighborhood search \cite{8} addresses the static DARP using three classes of neighborhoods. Feasibility testing of dial-a-ride problem under maximum waiting time and maximum ride time has been investigated in \cite{6}.

Baugh et al. \cite{2} were the first to employ simulated annealing to solve DARP. In 2003, Cordeau and Laporte \cite{TS} redefined the DARP formulation based on the real-life scenarios, and provided 20 benchmark instances. This formulation has been widely studied by the research community, and considered as the standard DARP, as detailed in \cite{DARP}. 

In this paper, we propose a multi-atomic annealing heuristic for DARP. The major contributions of the paper are summarized as follows:
\begin{itemize}
\item A multi-atomic annealing (MATA) heuristic for DARP is proposed, in which a new construction heuristic is used to generate initial solution.
\item Two new local search operators, \textit{burn} and \textit{reform}, are introduced to quickly explore the search space.
\item The proposed MATA algorithm is tested on various standard DARP benchmark instances \cite{TS} and analysed for convergence.
\end{itemize}

The remainder of the paper is organized as follows: In Section~\ref{sec:formulation}, we briefly discuss the standard DARP formulation. In Section~\ref{sec:mata}, we present the proposed multi-atomic annealing heuristic to solve DARP. In Section~\ref{sec:simulation}, we discuss the simulation results for the proposed algorithm. In Section~\ref{sec:conclusion}, we offer concluding remarks and ideas for future research.

\section{Problem Formulation}
\label{sec:formulation}
The primary objective of dial-a-ride problem is to minimize the overall transportation cost besides providing superior passenger comfort and safety. DARP is mathematically formulated as an optimization problem subjected to several constraints. The DARP formulation addressed in this paper was proposed by Cordeau and Laporte \cite{TS}. A homogeneous set of customers and fleet are considered. It is assumed that the travel times between vertices are known.

In DARP, $n$ customer requests are served using $m$ vehicles, where each request $i$ consists of time window either for departure or arrival vertex. Let $\mathbb{S} = \{x_1,x_2,\cdots\}$ denotes the search space, where $x_1,x_2,\cdots$ represent unique solutions. Every route for a vehicle $k$ starts and ends at the depot; the departure vertex $v_i$ and arrival vertex $v_{i+n}$ must belong to the same route; the arrival vertex $v_{i+n}$ is visited after departure vertex $v_i$. In addition, several other constraints need to be satisfied: (i) the load of vehicle $k$ cannot exceed the vehicle capacity $Q_k$ at any time; (ii) the total route duration of a vehicle $k$ cannot exceed predefined duration bound $T_k$; (iii) the ride time of any passenger cannot exceed the ride time bound $L$; the time window set by the customer must not be violated.

Therefore, four major constraints exist in dial-a-ride problem: load, duration, time window, and ride time constraints. Load constraint violation $q(x)$ occurs when the number of passenger in a vehicle $k$ exceeds its load limit $Q_k$; duration constraint violation $d(x)$ happens when a vehicle $k$ exceeds its duration limit $T_k$; time window constraint violation $w(x)$ appears when time at the start of service $B_i$ is later than $l_i$, the upper bound of time window; ride time constraint violation $t(x)$ occurs when a passenger is transported longer than ride time bound $L$. The constraints are defined as follows:
\begin{align}
q(x) &= \sum_{\forall k} \max(q_{k,\max}-Q_k,0) \\
d(x) &= \sum_{\forall k} \max(d_k-T_k,0) \\
w(x) &= \sum_{\forall i} \big[\max(B_i-l_i,0) + \max(B_{i+n}-l_{i+n},0)\big] \\
t(x) &= \sum_{\forall i} \max(L_i-L,0).
\label{eq:constraints}
\end{align}

In the next section, we present the proposed multi-atomic annealing algorithm to solve DARP.

\section{Multi-Atomic Annealing (MATA)}
\label{sec:mata}

Multi-atomic annealing (MATA) is inspired by the physical annealing process of solids, in which a solid is heated, and allowed to cool very slowly until it achieves a stable crystal lattice configuration with minimum lattice energy state. If the cooling schedule is sufficiently slow, the final configuration results in a solid with superior structural integrity.

The key algorithmic feature of MATA is that it provides means to escape local optima by melting a part of solid and allow the melted region to reform itself in a more stable configuration. At high temperature, a huge area of solid is melted, so more atoms are separated from the solid; at low temperature, a tiny area of solid is melted, so less atoms are separated from the solid. After melting, the structure is allowed to cool and reform into a more stable configuration with superior connecting bonds.

In literature, simulated annealing is another algorithm that mimics the annealing process. However, the concept of temperature defined in MATA is different from that of simulated annealing. In simulated annealing, the temperature is proportional to the energy of an atom involved in the diffusion process \cite{SA}. Whereas in MATA, temperature is proportional to the number of atoms separated and rearranged within the solid to improve the structure integrity.

It is evident that the temperature plays a vital role in determining the changes in structure. At high temperature, major changes to the solution are performed, allowing a wider exploration of search space; whereas at low temperature, minor changes to the solution are performed, allowing a deeper exploitation of the search space. Thus, a high initial temperature is preferred, so that a wider area of search space can be explored before a small area of search space can be exploited.

The notion of requests and solution quality in DARP is analogous to atoms and structural integrity of metals in metallurgy. To the best of our knowledge, this algorithm is proposed for the first time. The major challenge in this algorithm is to design a suitable cooling schedule, i.e. exploration and exploitation strategies, to obtain high quality solutions rapidly.

\subsection{Algorithm}
\label{sec:mata_alg}

\begin{algorithm}[!b]
\caption{\textit{MATA Algorithm}}
\label{al:mata_alg}
\begin{algorithmic}[1]
\Require $T_{\text{min}}$: minimum temperature; $T_{\text{max}}$: maximum temperature; $\lambda_T$: temperature drop rate; $\delta_{\text{max}}$: no improvement limit
\Ensure $x_{\text{best}}$: best solution found
%\Funct \textit{construct}(), \textit{burn}(), \textit{reform}()
\State $T \leftarrow T_{\text{max}}$, $\delta \gets 0$
\State $x_{\text{best}} \gets \varnothing$ where $f(x_{\text{best}}) = \inf$.
%\State Generate \textit{Sorted\_List}.
%\State $x_0 =$ \textit{Construction\_Heuristic}$(n,m)$.
%\State Construct initial solution $x_0$.
\State generate \textit{Sorted\_List}.
\State $x \gets$ \textit{construct}$($\textit{Sorted\_List}$)$.
\Repeat
    \State $x \gets$ \textit{burn}$(x,T,$\textit{Sorted\_List}$)$.
    \State $x \gets$ \textit{reform}$(x)$.
    \If {$x\in(\mathbb F\cap\mathbb C)$ \textbf{and} $f(x) < f(x_{\text{best}})$}
        \State $x_{\text{best}} \gets x$.
    \Else
        \State $\delta \gets \delta+1$.
    \EndIf
    \If {$\delta > \delta_{\text{max}}$}
        \State $x \gets x_{\text{best}}$, $\delta \gets 0$.
    \EndIf
    \State $T \gets T\times(1-\lambda_T).$
    \If {$T < T_{\text{min}}$}
        \State $T \gets$ random number in [$T_{\text{min}}$, $T_{\text{max}}$].
    \EndIf
\Until{termination criterion is fulfilled}
\State \Return $x_{\text{best}}$.			
\end{algorithmic}
\end{algorithm}

Multi-atomic annealing uses a construction heuristic to form an initial solution. The algorithm operates within a specified search space $\mathbb{S}$, where all solutions must be feasible without the need to be complete. A solution $x$ is in feasible space $\mathbb{F}$, when all constraints ($q(x), d(x), w(x),$ and $t(x)$) are not violated; a solution $x$ is in complete space $\mathbb{C}$, when all the requests are served. Search space, feasible space and complete space are defined as follows:
\begin{align}
\mathbb{F} &= \{x:q(x) = d(x) = w(x) = t(x) = 0\} \\
\mathbb{C} &= \{x:r(x) = n\} \\
\mathbb{S} &= \{x:x \in \mathbb{F}\},
\label{eq:f&c}
\end{align}
where $r(x)$ is the number of requests served in a solution. A solution $x$ has a cost $f(x)$ which is equivalent to the total travel time of all vehicles. Solutions are evaluated using the three-level neighborhood evaluation method \cite{ITS}, which minimizes all the constraint violations while performing scheduling.

Each iteration of the algorithm consists of three parts: In the first part, two local search operators, \textit{burn} and \textit{reform}, are used to alter the current solution to $x_B$ and $x_R$ respectively.
In the second part, the current solution is compared against the best known solution $x_{\text{best}}$. A solution $x$ is accepted as the best known solution, if and only if it is both feasible and complete ($x \in \mathbb{F} \cap\mathbb{C}$), besides having the least cost ($f(x)<f(x_{\text{best}})$). In the third part, the current temperature $T$ is updated with respect to the temperature drop rate $\lambda_T$, and a restart mechanism may be triggered. The optimization process is restarted when no better solution is found after
$\delta_{\text{max}}$ (no improvement limit) iterations. With the assumption that some better solutions are located near the current best known solution ($x_{\text{best}}$), the current solution $x$ is reset
to $x_{\text{best}}$ to facilitate efficient exploitation of search space around the best known solution.

The algorithm is repeated until the termination criterion is fulfilled. In this paper, we use run time as the termination criterion. Algorithm \ref{al:mata_alg} presents the structure of the multi-atomic annealing.

\begin{figure}[!htbp]
	\centering
	\subfloat[Unsorted requests.]{\includegraphics[width=0.24\textwidth]{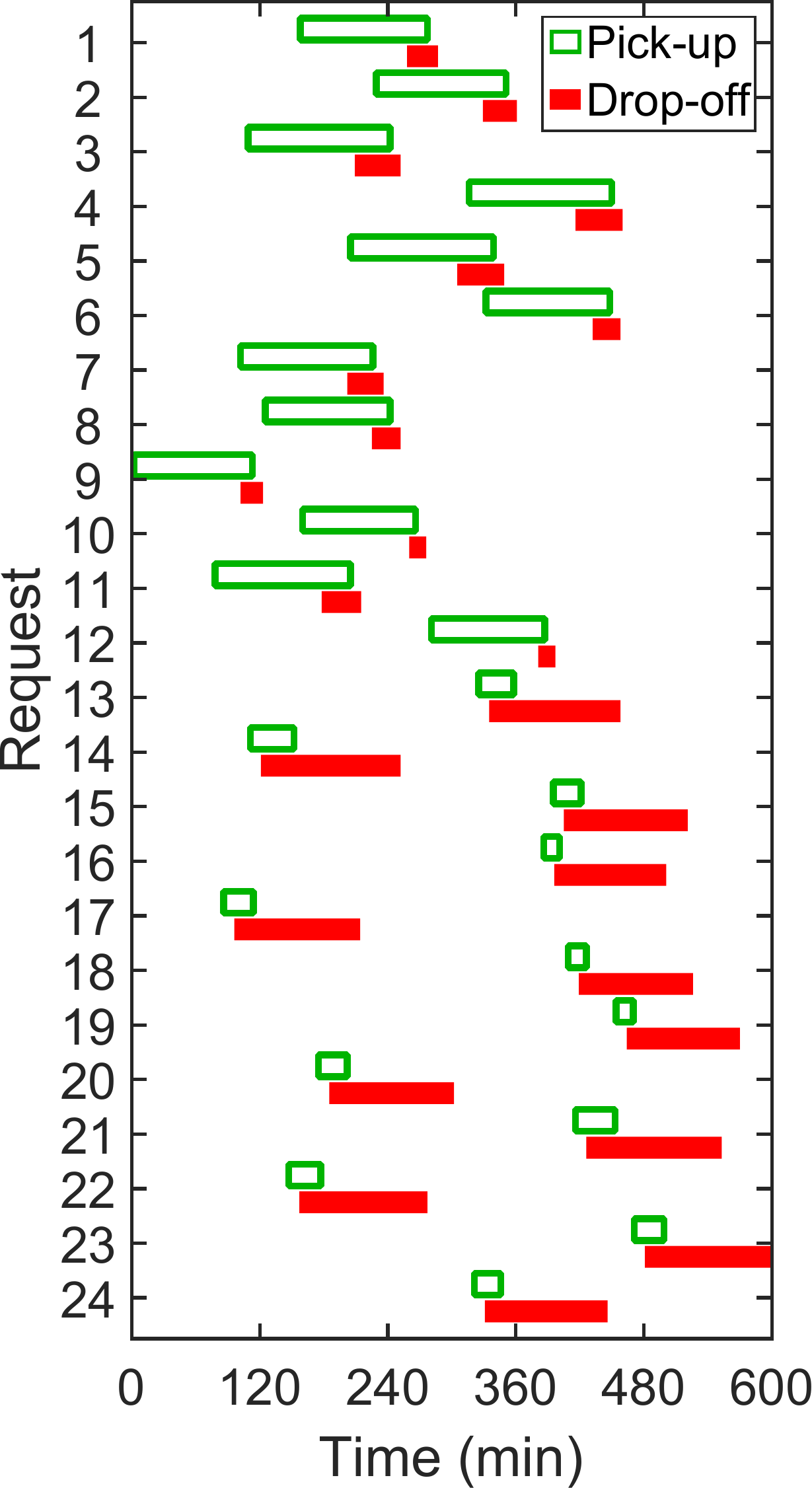}\label{fig:tw_unsorted}}
	\hfill
	\subfloat[Sorted requests.]{\includegraphics[width=0.24\textwidth]{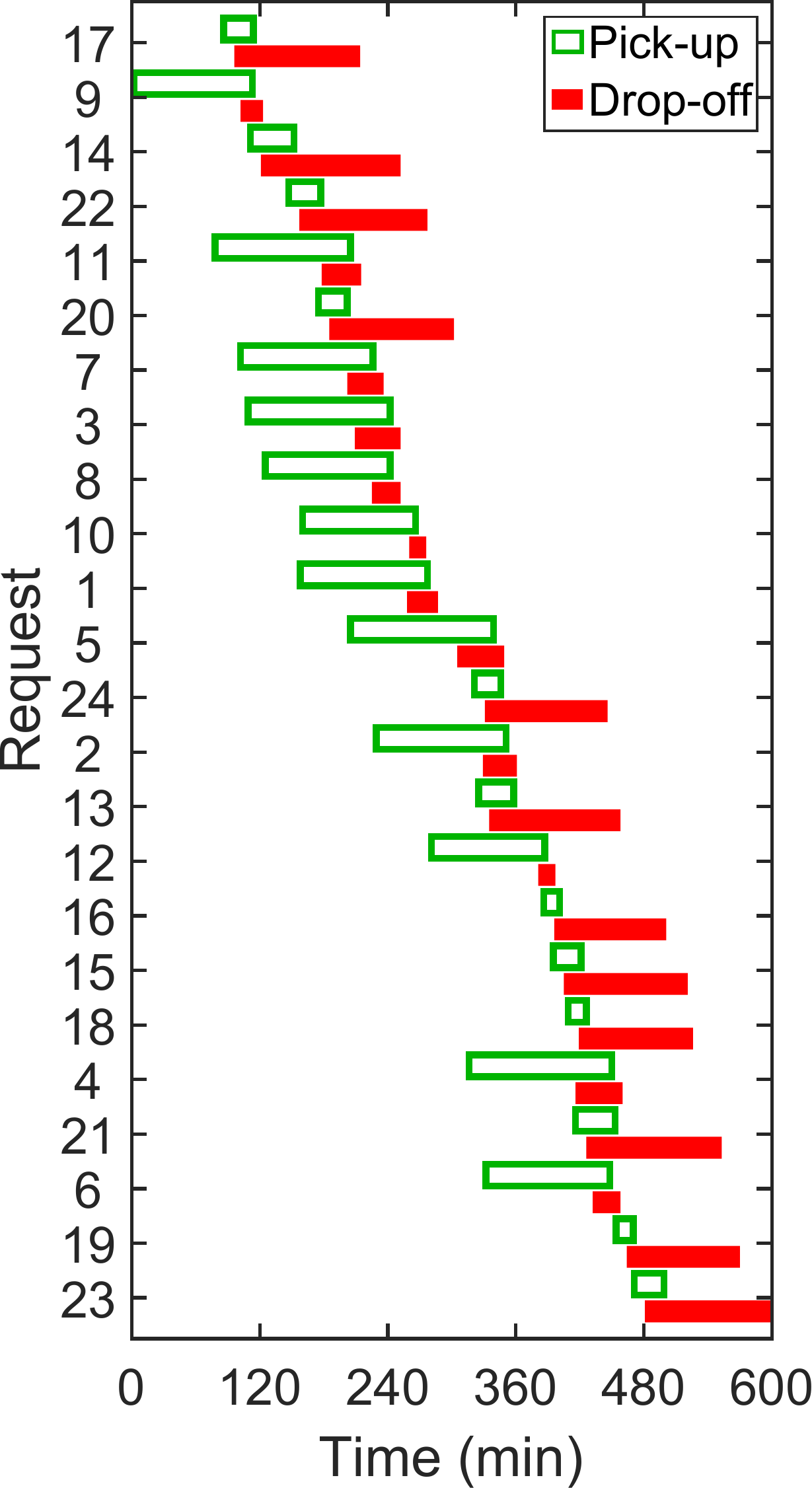}\label{fig:tw_sorted}}
    \caption{Request sequencing mechanism for R1a instance \cite{TS}.}
    \label{fig:req_sequencing}
\end{figure}

\subsection{Request Sequencing}
\label{sec:req_seq}

Two different request sequencing mechanisms are used: random and sorted. The random sequence labeled as  \textit{Random\_List}, is constructed by arranging all requests in random order. The sorted sequence labeled as \textit{Sorted\_List}, is constructed in two steps. In the first step, time window adjustment \cite{ITS} is performed. In the second step, requests are sorted in an ascending order according to the value of $l_i+e_{i+n}$, where $l_i$ is the upper bound of the time window for the departure vertex and $e_{i+n}$ is the lower bound of the time window for the arrival vertex. Figs. \ref{fig:tw_unsorted} and \ref{fig:tw_sorted} illustrate the request sequencing mechanism.

\subsection{Construction Heuristic}
\label{sec:const_heur}

Construction heuristic is used to form an initial solution by inserting all requests sequentially based on the \textit{Sorted\_List} (Section~\ref{sec:req_seq}). \textit{Sorted\_List} consists of requests arranged as per the time window, in which requests with time window approaching faster are kept ahead of others. Construction heuristic fetches requests from \textit{Sorted\_List} and inserts them into the solution at their best positions in a sequential order. During each insertion, a request is inserted into a route at a position with the lowest cost; if no feasible insertion is found, the request is not inserted. Two-step insertion, also known as neighborhood reduction \cite{TS}, is used throughout this paper. Although construction heuristic does not guarantee a complete initial solution (i.e. all requests are served), \textit{Sorted\_List} ensures that the constructed initial solution serves more requests before the optimization process begins. Algorithm \ref{al:ch} presents the construction heuristic.

\begin{algorithm}[!htbp]
\caption{Construction Heuristic: \textit{construct}()}
\label{al:ch}
\begin{algorithmic}[1]
\Require \textit{Sorted\_List}
\Ensure  $x_0$: initial solution
\State $x_0 \gets \varnothing$.
\ForAll{request $i$ \textbf{in} \textit{Sorted\_List}}
    \State insert request $i$ into $x_0$ at the best position.
\EndFor
\State \Return $x_0$.
\end{algorithmic}
\end{algorithm}

\subsection{Local Search Operators}
\label{sec:loc_search}

\begin{algorithm}[b]
\caption{Burn Operator: \textit{burn}()}
\label{al:burn}
\begin{algorithmic}[1]
\Require $x$: current solution; $T$: temperature; \textit{Sorted\_List}
\Ensure $x_B$: burned solution
%\Funct \textit{vertex\_set}().
\State $x_B \gets x$.
\State $R \gets$ random integer in $[1,T]$.
\State $i_{\text{start}} \gets$ random integer in $[1,n]$.
\State $i_{\text{end}} \gets \min(n,i_{\text{start}}+R)]$.
%\State Initialize $i_{\text{start}}$ = \{$i:~i\in$ [$1,n$] $\cap$ $\mathbb{Z}$\}.
%\State Initialize $i_{\text{end}}$ = \{$i:~i\in$ [$i_{\text{start}},\min(n,i_{\text{start}}+T)$] $\cap$ $\mathbb{Z}$\}.
\For {request $i \gets i_{\text{start}}$ \textbf{to} $i_{\text{end}}$ \textbf{in} \textit{Sorted\_List}}
    \State remove request $i$ from $x_B$.
\EndFor	
\State \Return $x_B$.		
\end{algorithmic}
\end{algorithm}

Two new local search operators, namely \textit{burn} and \textit{reform} are proposed in this algorithm. During \textit{burn} process, a band of requests is removed from the \textit{Sorted\_List}. As detailed in Section \ref{sec:mata}, the number of requests to be removed, $R$, is proportional to the current temperature $T$, where $R~\in~[1,T]$ is a random integer with equal chance of being selected during each execution of \textit{burn} operation. Algorithm \ref{al:burn} describes the \textit{burn} operation in detail.

During the \textit{reform} process, all requests are inserted in a random sequence \textit{Random\_List}, which results in an exhaustive search for all possible combinations of request insertions. If a request already exists in the solution, the request is removed before it is reinserted. During each insertion, a request is inserted into a route at a position with the lowest cost; if no feasible insertion is found, the request is not inserted. The process of solution reformation (\textit{reform}) is detailed in Algorithm \ref{al:reform}.

\begin{algorithm}
\caption{Reform Operator: \textit{reform}()}
\label{al:reform}
\begin{algorithmic}[1]
\Require $x$: current solution
\Ensure $x_R$: reformed solution
%\Funct \textit{vertex\_set}().
\State $x_R \gets x$.
\State generate \textit{Random\_List}.
\ForAll {request $i$ \textbf{in} \textit{Random\_List}}
    \If {request $i$ is served in $x_R$}
        \State remove request $i$ from $x_R$.
    \EndIf
    \State insert request $i$ into $x_R$ at the best position.
\EndFor	
\State \Return $x_R$.		
\end{algorithmic}
\end{algorithm}

\subsection{Restart Mechanism}
\label{sec:restart}
The optimization process is restarted when no better solution is found after $k_{max}$ iterations. With the assumption that some better solutions are located near the current best known solution, the current solution is reset such that $x \gets x_{best}$ to encourage more exploitation of search space near the best known solution.

We have tested the proposed MATA algorithm against other methods in the literature; the simulation results are presented in the next section.

\section{Simulation Results}
\label{sec:simulation}
The proposed multi-atomic annealing heuristic is implemented in C++. Simulations have been carried out on a computer running 2.1 GHz Intel Xeon E5-2620 v4 processor with 128 GB RAM. We consider the DARP benchmark instances R1a, R3a, R6a, R8a \cite{TS} to carry out the experiments. In these instances, the number of requests are 24, 72, 144, and 72, respectively; the fleet size is 3, 7, 13, and 6, respectively. The capacity of each vehicle is 6, the maximum ride time of a passenger is 90 min, and the maximum route duration is 480 min. We set the run time of 60 sec as the algorithm termination criterion. The maximum and minimum temperatures are set to $\frac{n}{2}$  and $\frac{m}{2}$ respectively, where $n$ is the number of requests and $m$ is the number of vehicles. The parameters $\delta_{max}$ and $\lambda_T$ are set to 30 and 0.01 respectively.

\begin{figure}[!t]
	\centering
	\includegraphics[width=0.48\textwidth]{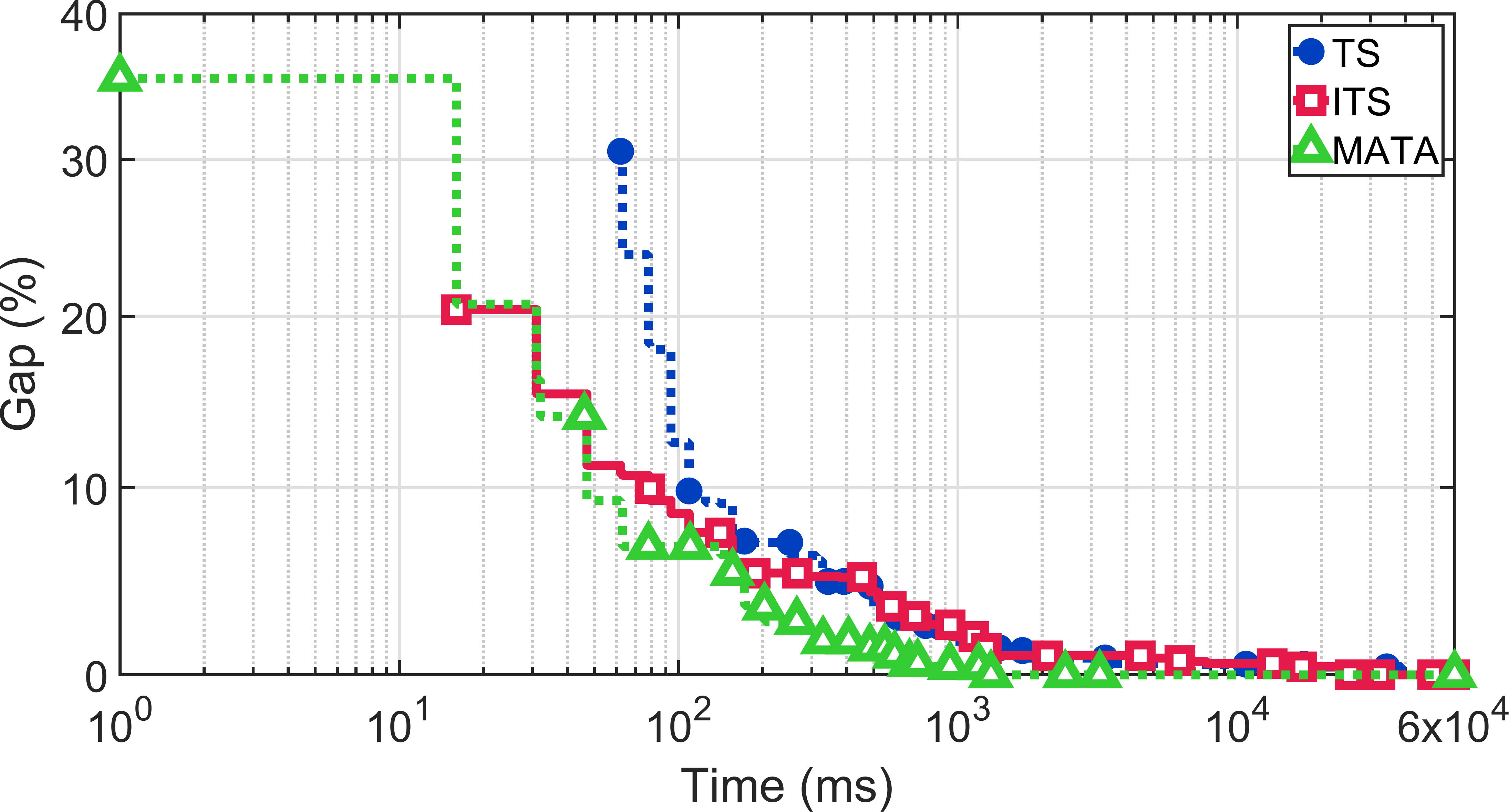}
    \caption{Comparison of MATA with other methods for R1a instance \cite{TS}.}
    \label{fig:p2_r1a}
\end{figure}
\begin{figure}[!t]
	\centering
	\includegraphics[width=0.48\textwidth]{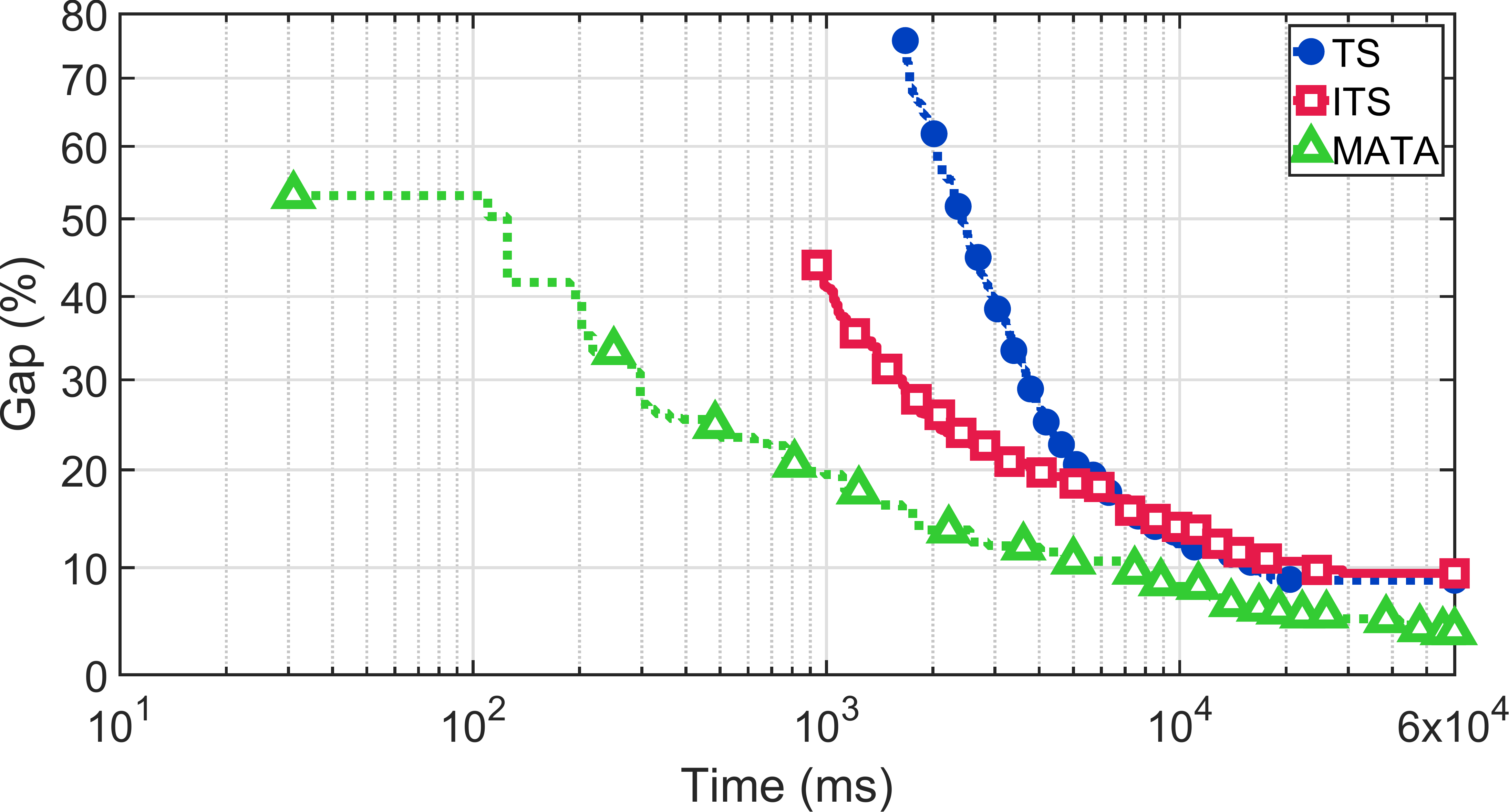}
    \caption{Comparison of MATA with other methods for R3a instance \cite{TS}.}
    \label{fig:p2_r3a}
\end{figure}
\begin{figure}[!t]
	\centering
	\includegraphics[width=0.48\textwidth]{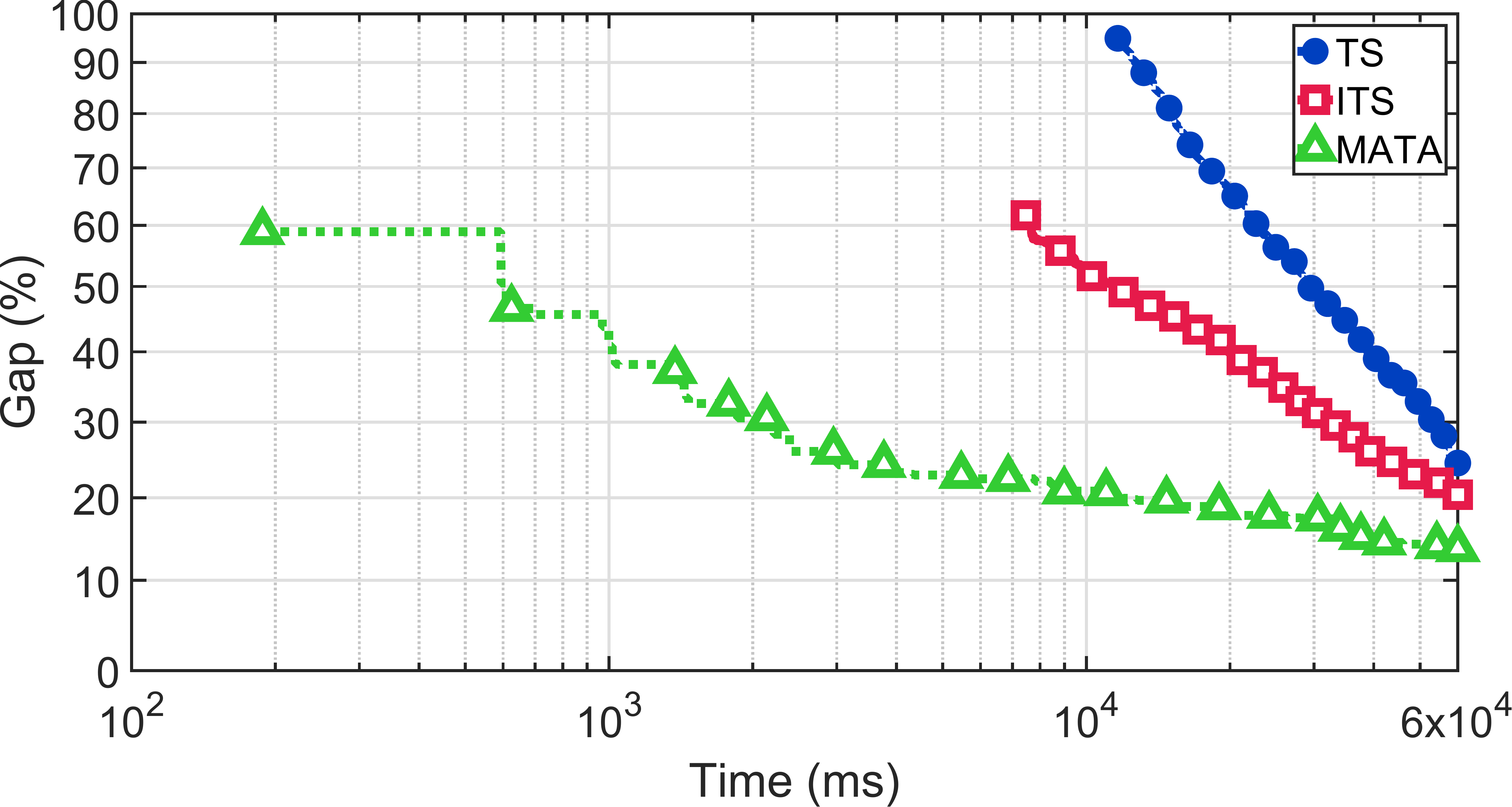}
    \caption{Comparison of MATA with other methods for R6a instance \cite{TS}.}
    \label{fig:p2_r6a}
\end{figure}
\begin{figure}[!t]
	\centering
	\includegraphics[width=0.48\textwidth]{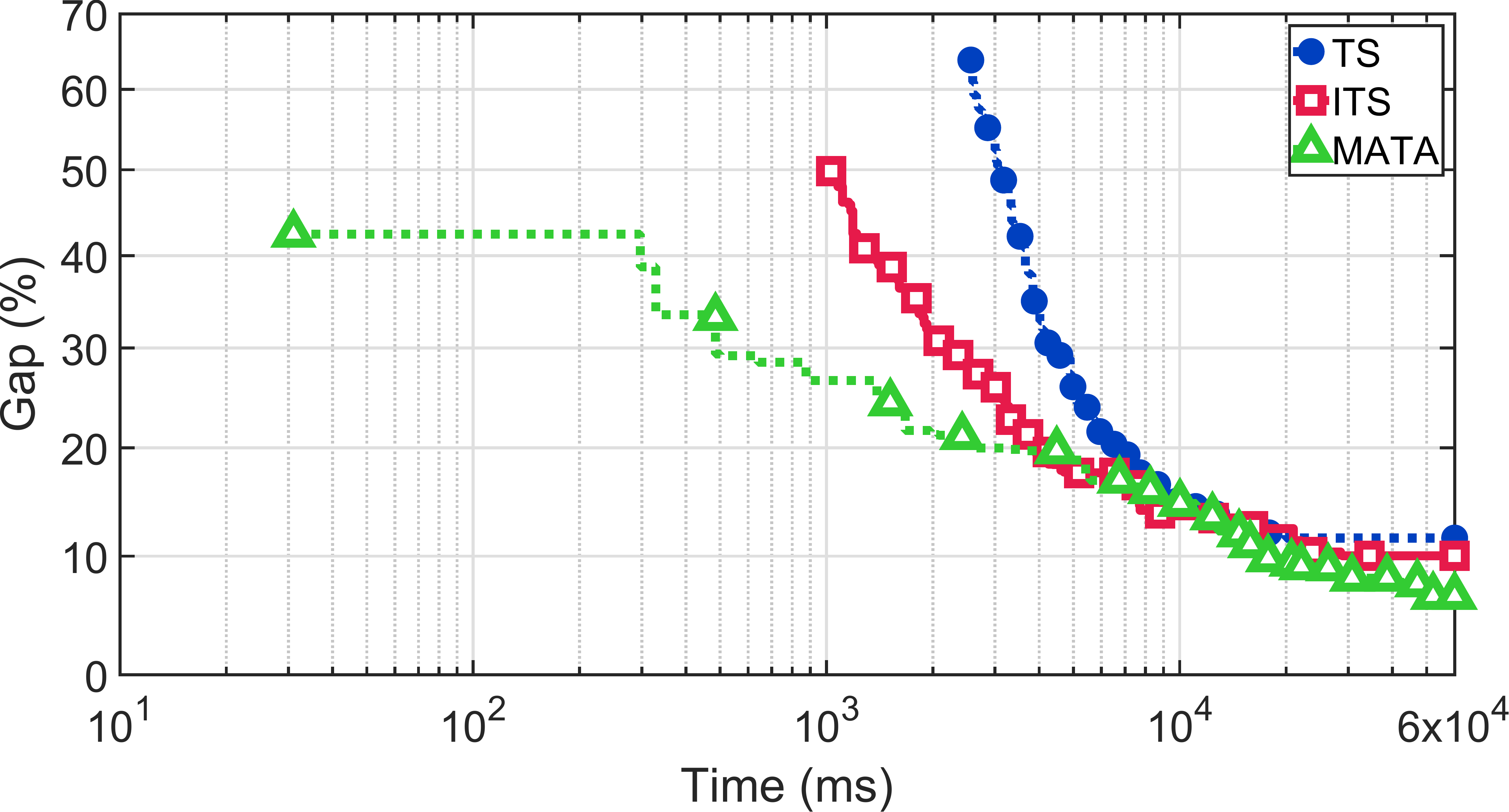}
    \caption{Comparison of MATA with other methods for R8a instance \cite{TS}.}
    \label{fig:p2_r8a}
\end{figure}

Figs. \ref{fig:p2_r1a} to \ref{fig:p2_r8a} illustrate the convergence of the proposed MATA when compared to tabu search (TS) \cite{TS} and improved tabu search (ITS) \cite{ITS}. The lines with triangle, circle and square markers in the plots correspond to the MATA, TS \cite{TS} and ITS \cite{ITS} algorithms respectively. The x-axis of the plots represents the simulation run time, and the y-axis corresponds to the gap (\%) as given by:
\begin{align}
\text{Gap (\%)}~=~\frac{\text{cost}~-~\textit{BKS}}{\textit{BKS}}\times100,
\label{eq:cost_difference}
\end{align}
where \textit{BKS} denotes the travel cost of the best known solution for a given instance. We plot the median of the gap recorded from five independent simulations ran for a duration of one minute. It is evident from these plots that the proposed MATA outperforms TS \cite{TS} and ITS \cite{ITS} algorithms in term of convergence. %Besides, the proposed algorithm also obtains global optimum solution for some of the instances.

\begin{table*}[!htbp]
	\centering
	\begin{tabular}{ccccccccccc}
	\toprule
	\multirow{2}{*}{Test Instance} & \multirow{2}{*}{\textit{BKS} \cite{Parragh}} & \multicolumn{3}{c}{Tabu Search (TS) \cite{TS}} & \multicolumn{3}{c}{Improved Tabu Search (ITS) \cite{ITS}} & \multicolumn{3}{c}{Multi-Atomic Annealing (MATA)} \\ 
	\cmidrule(lr){3-5}\cmidrule(lr){6-8}\cmidrule(lr){9-11}
	    &                      &  1 sec & 2 sec & 5 sec & 1 sec & 2 sec & 5 sec & 1 sec & 2 sec & 5 sec  \\ \midrule
	R1a	&190.02 & 193.26	&191.66	&191.11	&193.77	&191.88	&191.66	&190.84	&190.02	&190.02\\
	R3a	&532.00 &	-	&863.26	&644.51	&750.64	&670.69	&632.84	&635.77	&605.06	&588.60\\
	R6a	&785.26 &-	&-	&-	&-	&-	&-	&1105.87	&1025.38	&965.20\\
	R8a	&487.84 &-	&-	&614.78	&-	&642.78	&574.11	&617.85	&591.54	&579.57 \\
	\bottomrule
	\multicolumn{11}{l}{\em Note: `-' corresponds to the scenarios where there is no solution obtained within the mentioned time instance.} 
	\\ \\
	\end{tabular}
	\caption{Comparison of the travel cost at various time instances (1, 2 and 5 sec).}
	\label{tab:comparison1}
\end{table*}

\begin{table*}[!htbp]
	\centering
	\begin{tabular}{ccccccccccc}
	\toprule
    \multirow{2}{*}{Test Instance} & \multirow{2}{*}{\textit{BKS} \cite{Parragh}} & \multicolumn{3}{c}{Tabu Search (TS) \cite{TS}} & \multicolumn{3}{c}{Improved Tabu Search (ITS) \cite{ITS}} & \multicolumn{3}{c}{Multi-Atomic Annealing (MATA)} \\\cmidrule(lr){3-5}\cmidrule(lr){6-8}\cmidrule(lr){9-11}
                   &          & 15 sec & 30 sec & 60 sec  & 15 sec & 30 sec & 60 sec  & 15 sec & 30 sec & 60 sec  \\ \midrule
    R1a	& 190.02 &191.05	&190.79	&190.02	&191.11	&190.02	&190.02	&190.02	&190.02	&190.02 \\
    R3a	&532.00 &592.15	&578.66	&578.66	&593.06	&582.25	&582.25	&565.98	&559.20	&552.76 \\
    R6a	&785.26 &1416.25	&1171.76	&977.61	&1143.63	&1034.10	&945.55	&934.98	&922.40	&893.00 \\
    R8a	&487.84 &548.96	&544.45	&544.45	&549.66	&536.74	&536.74	&541.51	&528.25	&519.52\\
	\bottomrule
	\\
	\end{tabular}
	\caption{Comparison of the travel cost at various time instances (15, 30 and 60 sec).}
	\label{tab:comparison2}
\end{table*}

\begin{table*}[!htbp]
	\centering
	\begin{tabular}{ccccccccccc}
	\toprule
    \multirow{2}{*}{Test Instance} & \multirow{2}{*}{\textit{BKS} \cite{Parragh}} & \multicolumn{3}{c}{Tabu Search (TS) \cite{TS}} & \multicolumn{3}{c}{Improved Tabu Search (ITS) \cite{ITS}} & \multicolumn{3}{c}{Multi-Atomic Annealing (MATA)} \\\cmidrule(lr){3-5}\cmidrule(lr){6-8}\cmidrule(lr){9-11}
        &        & Cost & Gap (\%) & Time (ms) & Cost & Gap (\%) & Time (ms) & Cost & Gap (\%) & Time (ms) \\ \midrule
    R1a	& 190.02 & 248.05&	30.54&	62&	228.85	&20.43&	16 &257.48	&35.50	&\textbf{1}\\
    R3a	& 532.00 &  935.17&	75.78&	1672&	765.95&	43.98&	938 & 814.69	&53.14	&\textbf{31}\\
    R6a	& 785.26 &  1530.40&	94.89&	11642&	1269.81&	61.71&	7469 &1247.87	&58.91	&\textbf{188} \\
    R8a	& 487.84 &  799.18&	63.82&	2562&	731.00	&49.84&	1031 & 694.86	&42.44	&\textbf{31} \\
	\bottomrule
	\\
	\end{tabular}
	\caption{Comparison of first feasible solutions.}
	\label{tab:comparison3}
\end{table*}

In Tables \ref{tab:comparison1} and \ref{tab:comparison2}, we compare the median of the travel cost at various time instances \{1, 2, 5, 15, 30, 60\} sec for MATA, TS \cite{TS} and ITS \cite{ITS} algorithms. From these tables, it can be seen that the MATA ensures better solution quality when compared to the other algorithms. Based on \textit{BKS} listed in the tables, we can state that MATA attains near optimal solution in a shorter time for all the instances tested.

Table \ref{tab:comparison3} provides a comparison of the travel cost and execution time to attain first feasible solution for MATA, TS \cite{TS} and ITS \cite{ITS} on various DARP test instances. From Table \ref{tab:comparison3}, it is clear that MATA produces an initial feasible solution in less than a second, whereas the other algorithms require a considerable amount of time to achieve the same. As MATA obtains the first feasible solution much quicker, the quality of this solution might not be comparable to that of the other methods, which can be seen for instances R1a and R3a in Table \ref{tab:comparison3}. On average, MATA attains a first feasible solution 65.1\% faster than TS and 29.8\% faster than ITS; in 60 sec, MATA obtains a final solution that is 5.2\% better than TS and 3.9\% better than ITS.%Next section provides the concluding remarks for the paper.

\section{Conclusion}
\label{sec:conclusion}
In this paper, we proposed a novel multi-atomic annealing heuristic and applied it to solve static dial-a-ride problem. Many existing algorithms in the literature require long execution time to obtain a feasible solution. To overcome this limitation, we presented here two new local search operators (\textit{burn} and \textit{reform}, along with a construction heuristic, to provide a rapid feasible initial solution. Furthermore, MATA shows faster convergence when compared to other algorithms. We conducted various numerical experiments on standard DARP benchmark instances to validate these claims. Some possible directions for future work are: (i) refinement of the construction heuristic through the exploration of different sequencing mechanisms to enhance the quality of the first feasible solution, (ii) development of new local search operators to widen the search horizon, and (iii) parallel implementation of the MATA algorithm to further reduce the computation time.
%to address real-time scheduling and allocation problems. This greatly increases it's applicability in practice compared to other algorithms which are often more complex and require extensive computation resources. and state that the MATA heuristic not only improves the convergence to global optimum, but also finds a feasible initial solution in less than a second.

\section*{Acknowledgment}
The research was partially supported by the ST Engineering-NTU Corporate Lab 
through the National Research Foundation (NRF) corporate lab@university scheme. 

%\IEEEtriggeratref{7}
\bibliographystyle{IEEEtran}
\bibliography{ieee_conf}

\begin{thebibliography}{10}
\providecommand{\url}[1]{#1}
\csname url@samestyle\endcsname
\providecommand{\newblock}{\relax}
\providecommand{\bibinfo}[2]{#2}
\providecommand{\BIBentrySTDinterwordspacing}{\spaceskip=0pt\relax}
\providecommand{\BIBentryALTinterwordstretchfactor}{4}
\providecommand{\BIBentryALTinterwordspacing}{\spaceskip=\fontdimen2\font plus
\BIBentryALTinterwordstretchfactor\fontdimen3\font minus
  \fontdimen4\font\relax}
\providecommand{\BIBforeignlanguage}[2]{{%
\expandafter\ifx\csname l@#1\endcsname\relax
\typeout{** WARNING: IEEEtran.bst: No hyphenation pattern has been}%
\typeout{** loaded for the language `#1'. Using the pattern for}%
\typeout{** the default language instead.}%
\else
\language=\csname l@#1\endcsname
\fi
#2}}
\providecommand{\BIBdecl}{\relax}
\BIBdecl

\bibitem{VRP}
P.~Toth and D.~Vigo, \emph{Vehicle routing: problems, methods, and
  applications}.\hskip 1em plus 0.5em minus 0.4em\relax SIAM, 2014.

\bibitem{1}
N.~H. Wilson, J.~Sussman, H.-K. Wong, and T.~Higonnet, \emph{Scheduling
  algorithms for a dial-a-ride system}.\hskip 1em plus 0.5em minus 0.4em\relax
  Massachusetts Institute of Technology. Urban Systems Laboratory, 1971.

\bibitem{4}
J.-F. Cordeau, ``A branch-and-cut algorithm for the dial-a-ride problem,''
  \emph{Operations Research}, vol.~54, no.~3, pp. 573--586, 2006.

\bibitem{2}
J.~W. Baugh~Jr, G.~K.~R. Kakivaya, and J.~R. Stone, ``Intractability of the
  dial-a-ride problem and a multiobjective solution using simulated
  annealing,'' \emph{Engineering Optimization}, vol.~30, no.~2, pp. 91--123,
  1998.

\bibitem{TS}
J.-F. Cordeau and G.~Laporte, ``A tabu search heuristic for the static
  multi-vehicle dial-a-ride problem,'' \emph{Transportation Research Part B:
  Methodological}, vol.~37, no.~6, pp. 579--594, 2003.

\bibitem{7}
D.~Kirchler and R.~W. Calvo, ``A granular tabu search algorithm for the
  dial-a-ride problem,'' \emph{Transportation Research Part B: Methodological},
  vol.~56, pp. 120--135, 2013.

\bibitem{5}
M.~Chassaing, C.~Duhamel, and P.~Lacomme, ``An els-based approach with dynamic
  probabilities management in local search for the dial-a-ride problem,''
  \emph{Engineering Applications of Artificial Intelligence}, vol.~48, pp.
  119--133, 2016.

\bibitem{8}
S.~N. Parragh, K.~F. Doerner, and R.~F. Hartl, ``Variable neighborhood search
  for the dial-a-ride problem,'' \emph{Computers \& Operations Research},
  vol.~37, no.~6, pp. 1129--1138, 2010.

\bibitem{6}
M.~Firat and G.~J. Woeginger, ``Analysis of the dial-a-ride problem of hunsaker
  and savelsbergh,'' \emph{Operations Research Letters}, vol.~39, no.~1, pp.
  32--35, 2011.

\bibitem{DARP}
J.-F. Cordeau and G.~Laporte, ``The dial-a-ride problem: models and
  algorithms,'' \emph{Annals of Operations Research}, vol. 153, pp. 29--46,
  2007, https://doi.org/10.1007/s10479-007-0170-8.

\bibitem{SA}
S.~Kirkpatrick, C.~D. Gelatt, and M.~P. Vecchi, ``Optimization by simulated
  annealing,'' \emph{science}, vol. 220, no. 4598, pp. 671--680, 1983.

\bibitem{ITS}
S.~{Ho}, S.~C. {Nagavarapu}, R.~{Ramasamy Pandi}, and J.~{Dauwels}, ``{Improved
  Tabu Search Heuristic for Static Dial-A-Ride Problem},'' \emph{ArXiv
  e-prints, https://arxiv.org/abs/1801.09547}, 2018.

\bibitem{Parragh}
S.~N. Parragh and V.~Schmid, ``Hybrid column generation and large neighborhood
  search for the dial-a-ride problem,'' \emph{Computers \& Operations
  Research}, vol.~40, no.~1, pp. 490--497, 2013.

\end{thebibliography}

\end{document}